\documentclass[twocolumn,natbib]{svjour3}

\smartqed

\usepackage{graphicx}
\usepackage{mathptmx}

\usepackage{color,xcolor}
\usepackage{epsfig}
\usepackage{graphicx}

\usepackage{adjustbox}
\usepackage{array}
\usepackage{booktabs}
\usepackage{colortbl}
\usepackage{float,wrapfig}
\usepackage{hhline}
\usepackage{multirow}
\usepackage{subcaption} 

\usepackage{amsfonts}
\usepackage{mathtools}
\usepackage{bm}
\usepackage{nicefrac}

\usepackage{changepage}
\usepackage{extramarks}
\usepackage{fancyhdr}
\usepackage{lastpage}
\usepackage{setspace}
\usepackage{soul}
\usepackage{xspace}

\usepackage[pagebackref=true,breaklinks=true,colorlinks,citecolor=blue,bookmarks=false]{hyperref}
\usepackage{url}

\usepackage{algorithm, algorithmic}
\usepackage{enumerate}

\newcolumntype{L}[1]{>{\raggedright\let\newline\\\arraybackslash\hspace{0pt}}m{#1}}
\newcolumntype{C}[1]{>{\centering\let\newline\\\arraybackslash\hspace{0pt}}m{#1}}
\newcolumntype{R}[1]{>{\raggedleft\let\newline\\\arraybackslash\hspace{0pt}}m{#1}}

\newcommand{\sect}[1]{Section~\ref{#1}}

\newcommand{\fig}[1]{Figure~\ref{#1}}
\newcommand{\tbl}[1]{Table~\ref{#1}}

\newcommand{\ignorethis}[1]{}

\DeclareMathOperator*{\argmin}{arg\,min}

\makeatletter
\DeclareRobustCommand\onedot{\futurelet\@let@token\@onedot}
\def\@onedot{\ifx\@let@token.\else.\null\fi\xspace}

\def\eg{\emph{e.g}\onedot} 
\def\ie{\emph{i.e}\onedot}

\makeatother

\definecolor{MyDarkBlue}{rgb}{0,0.08,1}
\definecolor{MyDarkGreen}{rgb}{0.02,0.6,0.02}
\definecolor{MyDarkRed}{rgb}{0.8,0.02,0.02}
\definecolor{MyDarkOrange}{rgb}{0.40,0.2,0.02}
\definecolor{MyPurple}{RGB}{111,0,255}
\definecolor{MyRed}{rgb}{1.0,0.0,0.0}
\definecolor{MyGold}{rgb}{0.75,0.6,0.12}
\definecolor{MyDarkgray}{rgb}{0.66, 0.66, 0.66}

\newcommand{\myparagraph}[1]{\vspace{-3pt}\paragraph{#1}}

\def\model{\textbf{H}ardware-Aware \textbf{A}utomated \textbf{Q}uantization\xspace}
\def\modelshort{HAQ\xspace}

\journalname{International Journal of Computer Vision}

\begin{document}

\title{Hardware-Centric AutoML for Mixed-Precision Quantization}
\titlerunning{International Journal of Computer Vision}

\author{Kuan Wang$^*$ \and
        Zhijian Liu$^*$ \and
        Yujun Lin$^*$ \and
        Ji Lin \and
        Song Han
}

\authorrunning{International Journal of Computer Vision}

\institute{Kuan Wang$^*$ \at
           Massachusetts Institute of Technology \at
           \email{kuanwang@mit.edu}
           \and
           Zhijian Liu$^*$ \at
           Massachusetts Institute of Technology \at
           \email{zhijian@mit.edu}
           \and
           Yujun Lin$^*$ \at
           Massachusetts Institute of Technology \at
           \email{yujunlin@mit.edu}
           \and
           Ji Lin \at
           Massachusetts Institute of Technology \at
           \email{jilin@mit.edu}
           \and
           Song Han \at
           Massachusetts Institute of Technology \at
           \email{songhan@mit.edu}
           \and
           $*$ indicates equal contributions. 
}

\date{Received: date / Accepted: date}

\maketitle

\begin{abstract}

Model quantization is a widely used technique to compress and accelerate deep neural network (DNN) inference. Emergent DNN hardware accelerators begin to support \emph{mixed precision} (1-8 bits) to further improve the computation efficiency, which raises a great challenge to find the optimal bitwidth for each layer: it requires domain experts to explore the vast design space trading off accuracy, latency, energy, and model size, which is both time-consuming and usually sub-optimal. There are plenty of specialized hardware accelerators for neural networks, but little research has been done to design specialized neural networks optimized for a particular hardware accelerator. The latter is demanding given the much longer design cycle of silicon than neural nets. Conventional quantization algorithms ignore the different hardware architectures and quantize all the layers in a uniform way. In this paper, we introduce the \model (\modelshort) framework that automatically determines the quantization policy, and we take the hardware accelerator's feedback in the design loop. Rather than relying on proxy signals such as FLOPs and model size, we employ a hardware simulator to generate the direct feedback signals to the RL agent. Compared with conventional methods, our framework is fully automated and can provide specialized quantization policy for different nerual network and hardware architectures. The learned policy can transfer very well between different neural net architectures. Our framework effectively reduced the latency by \textbf{1.4-1.95$\times$} and the energy consumption by \textbf{1.9$\times$} with negligible loss of accuracy compared with the fixed bitwidth (8 bits) quantization. Our framework reveals that the optimal policies on different hardware architectures (\ie, edge and cloud architectures) under different resource constraints (\ie, latency, energy, and model size) are drastically different. We interpreted the implication of different quantization policies, which offer insights for both neural network architecture design and hardware architecture design.

\begin{figure}[t]
    \centering
    \includegraphics[width=\linewidth]{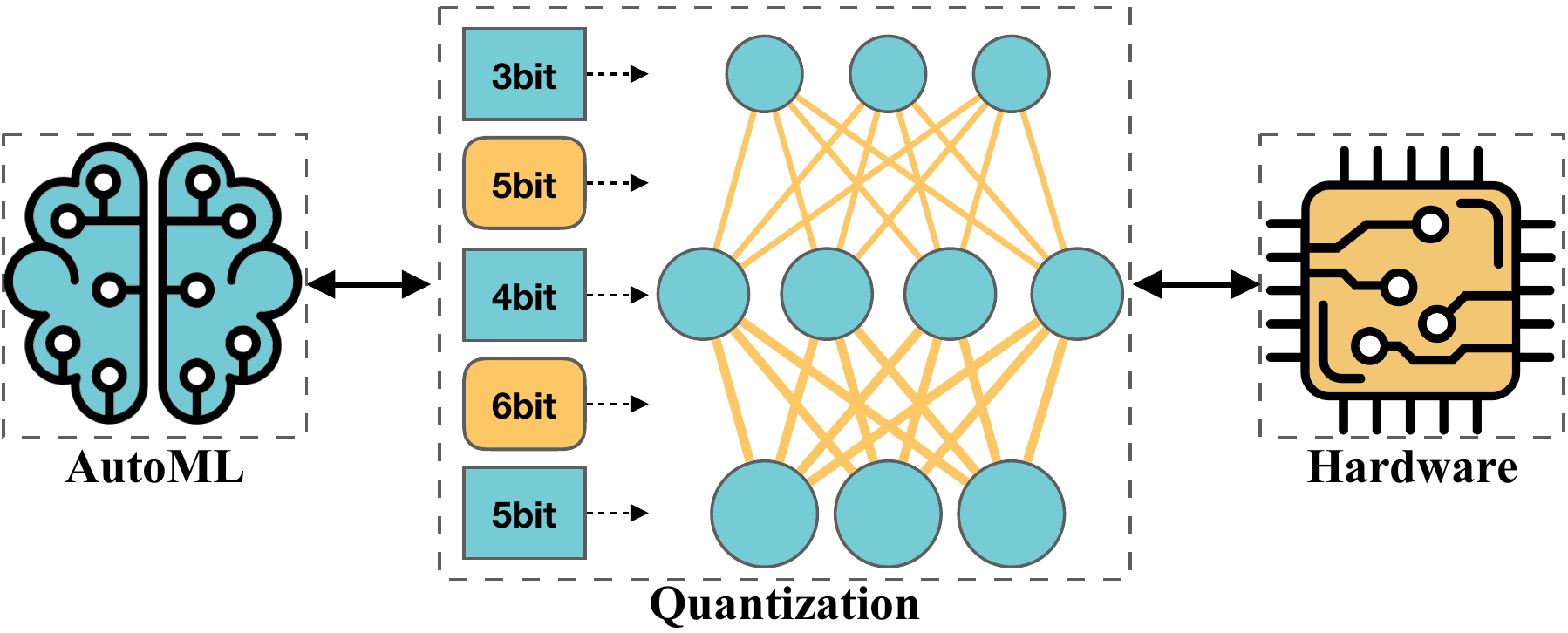}
    \caption{{Hardware-centric} (\emph{right}) {automated ML} (\emph{left}) for {mixed-precision quantization} (\emph{middle}).}
\end{figure}

\begin{figure}[t]
    \centering
    \includegraphics[width=\linewidth]{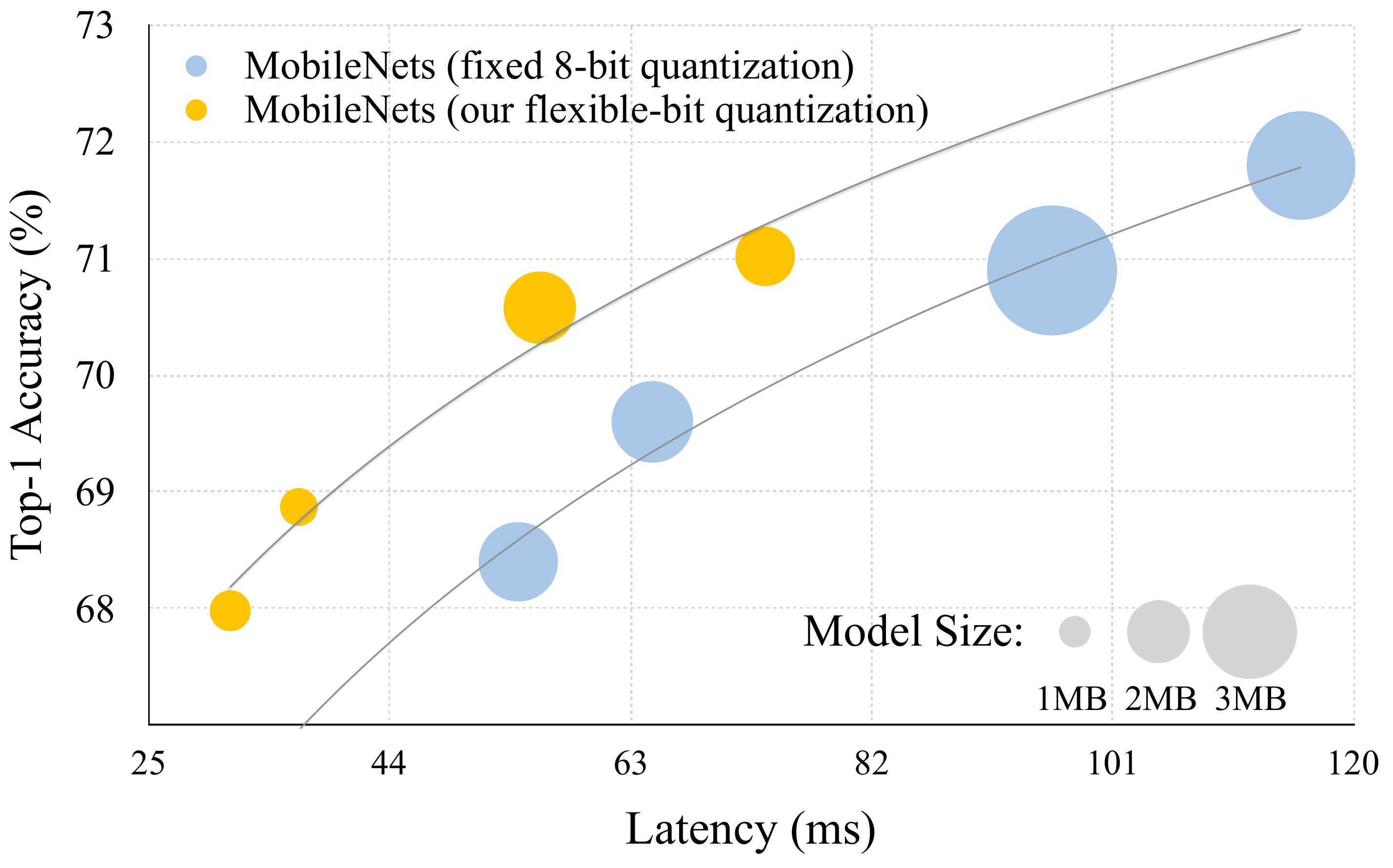}
    \caption{We need \emph{different} number of bits for different layers. We quantize MobileNets~\citep{Howard:2017wz} to different number of bits (both weights and activations), and it lies on a better pareto curve (yellow) than fixed-bit quantization (blue). This is because different layers have different redundancy and have different operation intensities (operations per byte) on the hardware, which advocates for using flexible bitwidths for different layers.}
    \label{fig:teaser}
\end{figure}


\end{abstract}

\section{Introduction}
\label{sec:intro}

In many real-time machine learning applications (such as robotics, autonomous driving, and mobile VR/AR), deep neural networks is strictly constrained by the latency, energy, and model size. In order to improve the hardware efficiency, many researchers have proposed to directly design efficient models~\citep{Sandler:2018wy,Howard:2017wz,Cai:2019ui} or to quantize the weights and activations to low precision~\citep{Han:2016uf,Zhu:2017wy}.

Conventional quantization methods use the same number of bits for all layers~\citep{Choi:2018uw,Jacob:2018ur}, but as different layers have different redundancy and behave differently on the hardware (computation bounded or memory bounded), it is necessary to use \emph{flexible} bitwidths for different layers (as shown in \fig{fig:teaser}). This flexibility was originally not supported by chip vendors until recently the hardware manufacturers started to implement this feature: Apple released the A12 Bionic chip that supports flexible bits for the neural network inference~\citep{apple}; NVIDIA recently introduced the Turing GPU architecture that supports 1-bit, 4-bit, 8-bit and 16-bit arithmetic operations~\citep{nvidia}; Imagination launched a flexible neural network IP that supports per-layer bitwidth adjustment for both weights and activations~\citep{imagination}. Besides industry, recently academia also works on the bit-level flexible hardware design: BISMO~\citep{umuroglu2018bismo} proposed the bit-serial multiplier to support multiplications of 1 to 8 bits; BitFusion~\citep{sharma2018bit} supports multiplications of 2, 4, 8 and 16 bits in a spatial manner.

\begin{table}[!t]
    \renewcommand*{\arraystretch}{1.4}
    \small\centering
    \begin{tabular}{lc|c|c}
        \toprule
        & \multicolumn{3}{c}{Inference latency on} \\
        & \textbf{HW1} & \textbf{HW2} & \textbf{HW3} \\
        \midrule
        Best Q. policy for \textbf{HW1} & \cellcolor{red!15}\textbf{16.29} ms & 85.24 ms & 117.44 ms \\
        Best Q. policy for \textbf{HW2} & 19.95 ms & \cellcolor{red!15}\textbf{64.29} ms & 108.64 ms \\
        Best Q. policy for \textbf{HW3} & 19.94 ms & 66.15 ms & \cellcolor{red!15}\textbf{99.68} ms \\
        \bottomrule
    \end{tabular}
    \caption{Inference latency of MobileNet-V1~\citep{Howard:2017wz} on three hardware architectures under different quantization policies. The quantization policy that is optimized for one hardware is not optimal for the other. This suggests we need a \emph{specialized} quantization solution for different hardware architectures. (HW1: BitFusion \citep{sharma2018bit}, HW2: BISMO \citep{umuroglu2018bismo} edge accelerator, HW3: BISMO cloud accelerator, batch = 16).}
    \label{tbl:teaser}
\end{table}

A very important missing part is, however, how to \textbf{determine the bitwidth of both weights and activations for each layer on different hardware accelerators}. This is a vast design space: with $M$ different neural network models, each with $N$ layers, on $H$ different hardware platforms, there are in total $O(H \times M \times 8^{2N})$ possible solutions (Here, we assume that the bitwidth is between 1 to 8 for both weights and activations). For a widely used ResNet-50~\citep{He:2015tt} model, the size of the search space is about $8^{100}$, which is even larger than the number of particles in the universe. Conventional methods require domain experts (with knowledge of both machine learning and hardware architecture) to explore the huge design space smartly with rule-based heuristics. For instance, we should retain more bits in the first layer which extracts low level features and in the last layer which computes the final outputs; also, we should use more bits in the convolution layers than in the fully-connected layers because empirically, the convolution layers are more sensitive. As the neural network becomes deeper, the exploration space increases exponentially, which makes it infeasible to rely on hand-crafted strategies. Therefore, these \emph{rule-based} quantization policies are usually sub-optimal, and they cannot generalize well from one model to another. In this paper, we would like to \emph{automate} this exploration process by a \emph{learning-based} framework.

Another challenge is how to measure the latency and the energy consumption of a given model on the hardware. A widely adopted approach~\citep{Howard:2017wz,Sandler:2018wy} is to rely on some proxy signals (\eg, FLOPs, number of memory references). However, as different hardware behaves very differently, the performance of a model on the hardware cannot always be accurately reflected by these proxy signals. Therefore, it is of great importance to directly \emph{involve the hardware architecture into the loop}. Also, as demonstrated in \tbl{tbl:teaser}, the quantization solution optimized on one hardware  might not be optimal on the other, which raises the demand for \emph{specialized} policies for different hardware architectures.

To this end, we propose the \model (\modelshort) framework that leverages reinforcement learning to automatically predict the quantization policy given the hardware's feedback. The RL agent decides the bitwidth of a given neural network in a layer-wise manner. For each layer, the agent receives the layer configuration and statistics as observation, and it then outputs the action which is the bitwidth of weights and activations. We then leverage the hardware accelerator as the environment to obtain the \emph{direct feedback from hardware} to guide the RL agent to satisfy the resource constraints. After all layers are quantized, we finetune the quantized model for one more epoch, and feed the validation accuracy after short-term retraining as the reward signal to our RL agent. During the exploration, we leverage the deep deterministic policy gradient (DDPG)~\citep{Lillicrap:2016ww} to supervise our RL agent. We studied the quantization policy on multiple hardware architectures: both cloud and edge neural network accelerators, with spatial or temporal multi-precision design.

The contribution of this paper has four aspects:
\begin{enumerate}
    \item \textbf{Automation}: We propose an automated framework for quantization, which does not require domain experts and rule-based heuristics. It frees the human labor from exploring the vast search space of choosing bitwidths.
    \item \textbf{Hardware-Aware}: Our framework integrates the hardware architecture into the loop so that it can directly reduce the latency, energy and storage on the target hardware instead of relying on some proxy signals.
    \item \textbf{Specialization}: For different hardware architectures, our framework can offer a specialized quantization policy that's exactly tailored for the hardware architecture.
    \item \textbf{Design Insights}: We interpreted the different quantization polices learned for different hardware architectures. Taking both computation and memory access into account, the interpretation offers insights on both neural network architecture and hardware architecture design.
\end{enumerate}

\section{Related Work}

\begin{figure*}[t]
    \centering
    \includegraphics[width=\textwidth]{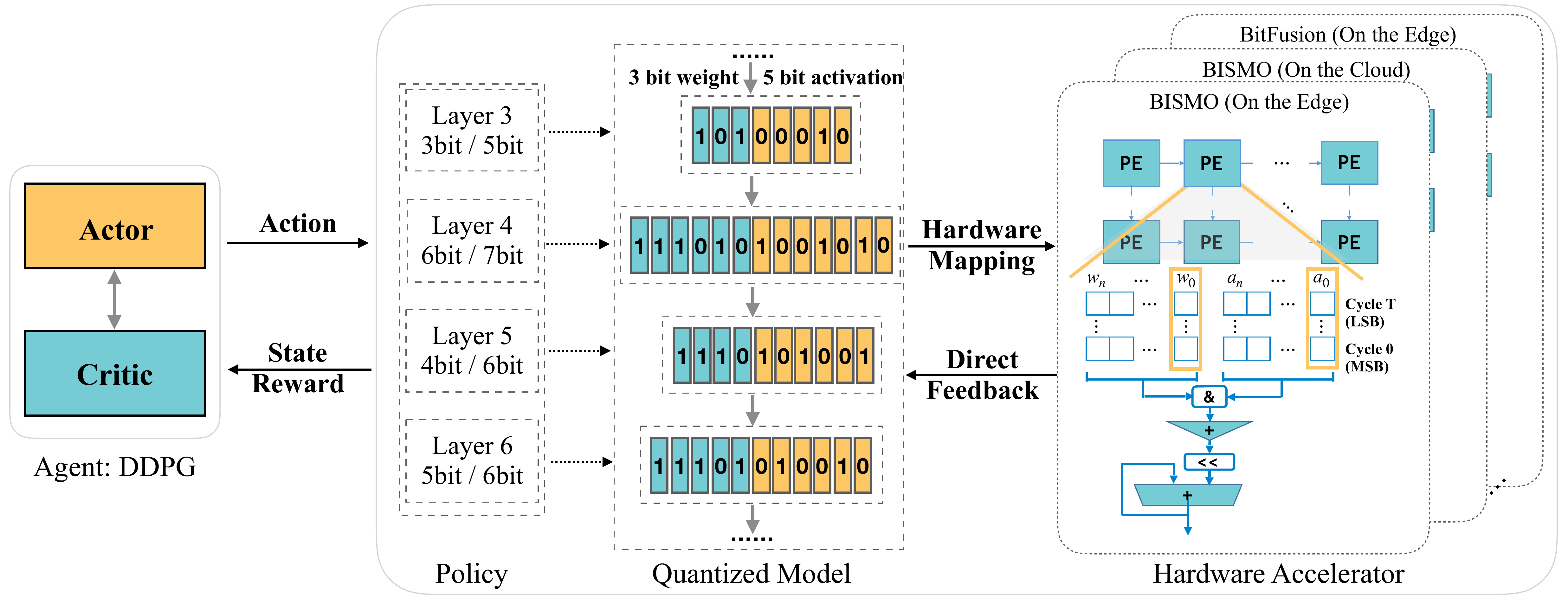}
    \caption{An overview of our \model (\modelshort) framework. We leverage the reinforcement learning to automatically search over the huge quantization design space with hardware in the loop. The agent propose an optimal bitwidth allocation policy given the amount of computation resources (\ie, latency, power, and model size). Our RL agent integrates the hardware accelerator into the exploration loop so that it can obtain the direct feedback from the hardware, instead of relying on indirect proxy signals.}
    \label{fig:overview}
\end{figure*}

\paragraph{Quantization.}

There have been extensive explorations on compressing and accelerating deep neural networks using quantization. \cite{Han:2016uf} quantized the network weights to reduce the model size by rule-based strategies: \eg, they used human heuristics to determine the bitwidths for convolution and fully-connected layers. \cite{Courbariaux:2016tm} binarized the network weights into $\{-1, +1\}$; \cite{Rastegari:2016tn} and \cite{zhou2018explicit} binarized each convolution filter into $\{-w, +w\}$; \cite{Zhu:2017wy} mapped the network weights into $\{-w_\text{N}, 0, +w_\text{P}\}$ using two bits; \cite{Zhou:2016wh} used one bit for network weights and two bits for activations; \cite{Jacob:2018ur} made use of 8-bit integers for both weights and activations. We refer the reader to the survey paper by \cite{Krishnamoorthi:2018wr} for a more detailed overview. These conventional quantization methods either simply assign the same number of bits to all layers or require domain experts to determine the bitwidths for different layers, while our framework automates this design process, and our \emph{learning-based} policy outperforms \emph{rule-based} strategies.

\myparagraph{Automated ML.}

Many researchers aimed to improve the performance of deep neural networks by searching the network architectures: \cite{Zoph:2017uo} proposed the Neural Architecture Search (NAS) to explore and design the transformable network building blocks, and their network architecture outperforms several human designed networks; \cite{Liu:2018tr} introduced the Progressive NAS to accelerate the architecture search by 5$\times$ using sequential model-based optimization; \cite{Pham:2018tl} introduced the Efficient NAS to speed up the exploration by 1000$\times$ using parameter sharing; \cite{Cai:2018wb} introduced the path-level network transformation to search the tree-structured architecture space effectively. Motivated by these AutoML frameworks, \cite{He:2018vj} leveraged the reinforcement learning to automatically prune the convolution channels. Our framework further explores the automated quantization for network weights and activations, and it takes the hardware architectures into consideration.

\myparagraph{Efficient Models.}

To facilitate the efficient deployment, researchers designed hardware-friendly approaches to slim neural network models. For instance, the coarse-grained channel pruning methods~\citep{he2017channel, liu2017learning} prune away the entire channel of convolution kernels to achieve speedup. 
Recently, researchers have explicitly optimized for various aspects of hardware properties, including the inference latency and energy: \cite{yang2016designing} proposed the energy-aware pruning to directly optimize the energy consumption of neural networks; \cite{yang2018netadapt} reduced the inference time of neural networks on the mobile devices through a lookup table.
Nevertheless, these methods are still rule-based and mostly focus on pruning. Our framework automates the quantization process by taking hardware-specific metric as direct rewards using a learning based method.
\section{Approach}

The overview of our proposed framework is in \fig{fig:overview}. We model the quantization task as a reinforcement learning problem. We used the actor critic model with DDPG agent to give action: bits for each layer. We collect hardware counters, together with accuracy as direct rewards to search the optimal quantization policy for each layer. We have three hardware environments that covers edge and cloud, spatial and temporal architectures for multi-precision accelerator. Below describes the details of the RL formulation.

\subsection{Observation (State Space)}

Our agent processes the neural network in a layer-wise manner. For each layer, our agent takes two steps: one for weights, and one for activations. In this paper, we introduce a ten-dimensional feature vector ${O}_k$ as our observation:

\vspace{4pt}\noindent
If the $k$\textsuperscript{th} layer is a convolution layer, the state ${O}_k$ is 
\begin{equation}
    O_k = (k, c_\text{in}, c_\text{out}, s_\text{kernel}, s_\text{stride}, s_\text{feat}, n_\text{params}, i_\text{dw}, i_\text{w/a}, a_{k - 1}),
\end{equation}
where $k$ is the layer index, $c_\text{in}$ is \#input channels, $c_\text{out}$ is \#output channels, $s_\text{kernel}$ is kernel size, $s_\text{stride}$ is the stride, $s_\text{feat}$ is the input feature map size, $n_\text{params}$ is \#parameters, $i_\text{dw}$ is a binary indicator for depthwise convolution, $i_\text{w/a}$ is a binary indicator for weight/activation, and $a_{k - 1}$ is the action from the last time step.

\vspace{4pt}\noindent
If the $k$\textsuperscript{th} layer is a fully-connected layer, the state ${O}_k$ is
\begin{equation}
    {O}_k = (k, h_\text{in}, h_\text{out}, 1, 0, s_\text{feat}, n_\text{params}, 0, i_\text{w/a}, a_{k-1}),
\end{equation}
where $k$ is the layer index, $h_\text{in}$ is \#input hidden units, $h_\text{out}$ is \#output hidden units, $s_\text{feat}$ is the size of input feature vector, $n_\text{params}$ is \#parameters, $i_\text{w/a}$ is a binary indicator for weight/ activation, and $a_{k - 1}$ is the action from the last step.

For each dimension in the observation vector ${O}_k$, we normalize it into $[0, 1]$ to make them in the same scale.

\subsection{Action Space}
\label{sect:approach:action_space}

We use a \emph{continuous} action space to determine the bitwidth. The reason that we do not use a \emph{discrete} action space is because it loses the relative order: \eg, 2-bit quantization is more aggressive than 4-bit and even more than 8-bit. At the $k$\textsuperscript{th} time step, we take the continuous action $a_k$ (which is in the range of $[0, 1]$), and round it into the discrete bitwidth value $b_k$:
\begin{equation}
    b_k = \text{round} (b_\text{min} - 0.5 + a_k \times (b_\text{max} - b_\text{min} + 1)),
\end{equation}
where $b_\text{min}$ and $b_\text{max}$ denote the min and max bitwidth (in our experiments, we set $b_\text{min}$ to $2$ and $b_\text{max}$ to $8$).

\myparagraph{Resource Constraints.}

In real-world applications, we have limited computation budgets (\ie, latency, energy, and model size). We would like to find the quantization policy with the best performance given the constraint.

We encourage our agent to meet the computation budget by limiting the action space. After our RL agent gives actions $\{a_k\}$ to all layers, we measure the amount of resources that will be used by the quantized model. The feedback is directly obtained from the hardware accelerator, which we will discuss in \sect{sect:hardware_accelerator}. If the current policy exceeds our resource budget (on latency, energy or model size), we will sequentially decrease the bitwidth of each layer until the constraint is finally satisfied.

\subsection{Direct Feedback from Hardware Accelerators}
\label{sect:hardware_accelerator}

An intuitive feedback to our RL agent can be FLOPs or the model size. However, as these proxy signals are indirect, they are not equal to the performance (\ie, latency, energy consumption) on the hardware.  Cache locality, number of kernel calls, memory bandwidth all matters.  Proxy feedback can not model these hardware functionality to find the specialized strategies (see \tbl{tbl:teaser}).

Instead, we use direct latency and energy feedback from hardware accelerators to optimize the performance. In simulators, the latency is approximated as the sum of the computation time, the stall caused by the memory access and some other overheads:
\begin{equation}
    T = T_{\text{computation}} + T_{\text{stall}} + T_{\text{overhead}},
\end{equation}
and the energy consumption is modeled as the sum of the logic circuits and memory:
\begin{equation}
\begin{aligned}
   E = \; & E_{\text{memory access per bit}} \times S_{\text{total memory access size}} \\
   &+ P_{\text{dynamic}} \times T_{\text{execution}}.
\end{aligned}
\end{equation}

The direct feedback from the hardware simulator is very crucial as it enables our RL agent to determine the bitwidth allocation policy from the subtle differences between different layers: \eg, the vanilla convolution has more data reuse and better locality, while the depthwise convolution~\citep{Chollet:2017vb} has less reuse and worse locality, which makes it memory bounded.

\subsection{Quantization}

We linearly quantize the weights and activations of each layer using the action $a_k$ given by our RL agent, as linearly quantized model only need fixed point arithmetic unit which is more efficient to implement on the hardware than the $k$-means quantization.

Specifically, for each weight value $w$ in the $k$\textsuperscript{th} layer, we first truncate it into the range of $[-c, c]$, and we then quantize it linearly into $a_k$ bits:
\begin{equation}
    \text{quantize}(w, a_k, c) = \text{round}(\text{clamp}(w, c) /s) \times s,
\end{equation}
where $\text{clamp}(\cdot, x)$ is to truncate the values into $[-x, x]$, and the scaling factor $s$ is defined as
\begin{equation}
    s = c / (2^{a_k - 1} - 1).
\end{equation}
In this paper, we choose the value of $c$ by finding the optimal value $x$ that minimizes the KL-divergence between the original weight distribution ${W}_k$ and the quantized weight distribution $\text{quantize}({W}_k, a_k, x)$:
\begin{equation}
    c = \argmin_x D_\text{KL} ({W}_k \mid\mid \text{quantize}({W}_k, a_k, x)),
\end{equation}
where $D_\text{KL}(\cdot \mid\mid \cdot)$ is the KL-divergence that characterizes the distance between two distributions. As for activations, we quantize the values similarly except that we truncate them into the range of $[0, c]$, not $[-c, c]$  since the activation values (which are the outputs of the ReLU layers) are non-negative. This calibration based on KL-divergence enables us to make use of the pretrained models rather than training the models from scratch so that it can reduce the training time significantly. As for the overhead, we only use 64 images to calibrate once at the beginning, which is negligible compared to the whole training.

\subsection{Reward Signal}

After quantization, we retrain the quantized model for one more epoch to recover the performance. As we impose the resource constraints by limiting the action space, we define our reward function $R$ to be only related to the accuracy:
\begin{equation}
    R = \lambda \times (\text{accuracy}_\text{quant} - \text{accuracy}_\text{origin}),
\end{equation}
where $\text{accuracy}_\text{origin}$ is the top-1 classification accuracy of the full-precision model on the training set, $\text{accuracy}_\text{quant}$ is the top-1 classification accuracy of the quantized model after finetuning, and $\lambda$ is a scaling factor which is set to $0.1$ in our experiments.

\subsection{Agent}

In our environment, one step means that our agent makes an action to decide the number of bits assigned to the weights or activations of a specific layer, while one episode is composed of multiple steps, where our RL agent makes actions to all layers. As for our RL agent, we leverage the deep deterministic policy gradient (DDPG)~\citep{Lillicrap:2016ww}, which is an off-policy actor-critic algorithm for continuous control problem. We apply a variant form of the Bellman's Equation, where each transition in an episode is defined as
\begin{equation}
    T_k = (O_k, a_k, R, O_{k+1}).
\end{equation}
During exploration, the ${Q}$-function is computed as
\begin{equation}
    \hat{Q}_k = R_k - B + \gamma \times Q(O_{k+1}, w(O_{k+1}) \mid \theta^Q),
\end{equation}
and the gradient signal can be approximated using
\begin{equation}
    L = \frac{1}{N_\text{s}} \sum_{k=1}^{N_\text{s}} (\hat{Q}_k - Q(O_k, a_k \mid \theta^Q))^2, \\
\end{equation}
where $N_\text{s}$ denotes the number of steps in this episode, and the baseline $B$ is defined as an exponential moving average of all previous rewards in order to reduce the variance of the gradient estimation. The discount factor $\gamma$ is set to 1 since we assume that the action made for each layer should contribute equally to the final result. Moreover, as the number of steps is always finite (bounded by the number of layers), the sum of the rewards will not explode.

\subsection{Implementation Details}

In this section, we present some implementation details about the RL agent, exploration and finetuning quantized models.

\myparagraph{Agent.}

The DDPG agent consists of an actor network and a critic network. Both follow the same network architecture: each network has 3 fully-connected layers with the hidden size of $[400, 300]$. For the actor network, the input is the state vector, and the output action is normalized to $[0, 1]$ by the sigmoid function; while for the critic network, the input is a vector concatenated by state and its corresponding action produced by actor.

\myparagraph{Exploration.}

Optimization of the DDPG agent is carried out using ADAM~\citep{Kingma:2015us} with $\beta_1 = 0.9$ and $\beta_2 = 0.999$. We use a fixed learning rate of $10^{-4}$ for the actor network and $10^{-3}$ for the critic network.
During exploration, we employ the following stochastic process of the noise:
\begin{equation}
    w'({O}_k) \sim N_\text{trunc} (w({O}_k \mid \theta^w_k), \sigma^2, 0, 1),
\end{equation}
where $N_\text{trunc}(\mu, \sigma, a, b)$ is the truncated normal distribution, and $w$ is the model weights. The noise $\sigma$ is initialized as $0.5$, and after each episode, the noise is decayed exponentially with a decay rate of $0.99$.  

\myparagraph{Finetuning.}

During exploration, we finetune the quantized model for one epoch to help recover the performance (using SGD with a fixed learning rate of $10^{-3}$ and momentum of $0.9$). We randomly select 100 categories from ImageNet~\citep{Deng:2009td} to accelerate the model finetuning during exploration. After exploration, we quantize the model with our best policy and finetune it on the full dataset.

\section{Experiments}
\label{sect:exp}

We conduct extensive experiments to demonstrate the consistent effectiveness of our framework for multiple objectives: \emph{latency}, \emph{energy}, \emph{model size}, and \emph{accuracy}.

\myparagraph{Datasets and Models.}

Our experiments are performed on the ImageNet~\citep{Deng:2009td} dataset. As our focus is on more efficient models, we extensively study the quantization of MobileNet-V1~\citep{Howard:2017wz} and MobileNet-V2~\citep{Sandler:2018wy}.  Both MobileNets are inspired from the depthwise separable convolutions~\citep{Chollet:2017vb} and replace the regular convolutions with the \emph{pointwise} and \emph{depthwise} convolutions: MobileNet-V1 stacks multiple ``\emph{depthwise} -- \emph{pointwise}" blocks repeatedly; while MobileNet-V2 uses the ``\emph{pointwise} -- \emph{depthwise} -- \emph{pointwise}" blocks as its basic building primitives.

\begin{table}[t]
    \renewcommand*{\arraystretch}{1.2}
    \setlength{\tabcolsep}{4pt}
    \small\centering
    \begin{tabular}{lcccccc} 
        \toprule
        & Hardware & Batch & PE Array & AXI port & Block RAM \\
        \midrule
        Edge & Zynq-7020 & 1 & 8$\times$8 & 4$\times$64b & 140$\times$36Kb \\
        Cloud & VU9P & 16 & 16$\times$16 & 4$\times$256b & 2160$\times$36Kb \\
        \bottomrule
    \end{tabular}
    \vspace{-6pt}
    \caption{Configurations of edge and cloud accelerators.}
    \label{tbl:edge_cloud}
\end{table}

\subsection{Latency-Constrained Quantization}

We first evaluate our framework under latency constraints on two representative hardware architectures: spatial and temporal architectures for multi-precision CNN:

\myparagraph{Temporal Architecture.}

Bit-Serial Matrix Multiplication Overlay (BISMO) proposed by \cite{umuroglu2018bismo} is a classic temporal design of neural network accelerator on FPGA. It introduces bit-serial multipliers which are fed with one-bit digits from 256 weights and corresponding activations in parallel at one time and accumulates their partial products by shifting over time.
 
\myparagraph{Spatial Architecture.}

BitFusion architecture proposed by \cite{sharma2018bit} is a state-of-the-art spatial ASIC design for neural network accelerator. It employs a 2D systolic array of Fusion Units which spatially sum the shifted partial products of two-bit elements from weights and activations. 

\begin{table*}[!t]
    \renewcommand*{\arraystretch}{1.2}
    \setlength{\tabcolsep}{4pt}
    \small\centering
    \begin{tabular}{lccccccccccccccc} 
        \toprule
        & & \multicolumn{6}{c}{Edge Accelerator} & \multicolumn{6}{c}{Cloud Accelerator}  \\
        \cmidrule(lr){3-8}\cmidrule(lr){9-14}

        & & \multicolumn{3}{c}{MobileNet-V1} & \multicolumn{3}{c}{MobileNet-V2} & \multicolumn{3}{c}{MobileNet-V1} & \multicolumn{3}{c}{MobileNet-V2} \\
        \cmidrule(lr){3-5}\cmidrule(lr){6-8}\cmidrule(lr){9-11}\cmidrule(lr){12-14}
        
        & Bitwidths & Acc.-1 & Acc.-5 & Latency & Acc.-1 & Acc.-5 & Latency & Acc.-1 & Acc.-5 & Latency & Acc.-1 & Acc.-5 & Latency \\
        \midrule
        PACT & 4 bits & 62.44 & 84.19 & 45.45 ms & 61.39 & 83.72 & 52.15 ms & 62.44 & 84.19 & 57.49 ms & 61.39 & 83.72 & 74.46 ms \\
        Ours & \emph{flexible} & \textbf{67.40} & \textbf{87.90} & 45.51 ms & \textbf{66.99} & \textbf{87.33} & 52.12 ms & \textbf{65.33} & \textbf{86.60} & 57.40 ms & \textbf{67.01} & \textbf{87.46} & 73.97 ms \\
        \midrule
        PACT & 5 bits & 67.00 & 87.65 & 57.75 ms & 68.84 & 88.58 & 66.94 ms & 67.00 & 87.65 & 77.52 ms & 68.84 & 88.58 & 99.43 ms \\
        Ours & \emph{flexible} & \cellcolor{red!15}\textbf{70.58} & \cellcolor{red!15}\textbf{89.77} & \cellcolor{red!15}57.70 ms & \textbf{70.90} & \textbf{89.91} & 66.92 ms & \cellcolor{blue!15}\textbf{69.97} & \cellcolor{blue!15}\textbf{89.37} & \cellcolor{blue!15}77.49 ms & \textbf{69.45} &\textbf{88.94} & 99.07 ms  \\
        \midrule
        PACT & 6 bits & 70.46 & 89.59 & 70.67 ms & 71.25 & 90.00 & 82.49 ms & 70.46 & 89.59 & 99.86 ms & 71.25 & 90.00 & 127.07 ms \\
        Ours & \emph{flexible} & \textbf{71.20} & \textbf{90.19} & 70.35 ms & \cellcolor{violet!15}\textbf{71.89} & \cellcolor{violet!15}\textbf{90.36} & \cellcolor{violet!15}82.34 ms & \textbf{71.20} & \textbf{90.08} & 99.66 ms & \cellcolor{olive!15}\textbf{71.85} & \cellcolor{olive!15}\textbf{90.24} & \cellcolor{olive!15}127.03 ms  \\
        \midrule
        Original & 8 bits & \cellcolor{red!15}70.82 & \cellcolor{red!15}89.85 & \cellcolor{red!15}96.20 ms & \cellcolor{violet!15}71.81 & \cellcolor{violet!15}90.25 & \cellcolor{violet!15}115.84 ms & \cellcolor{blue!15}70.82 & \cellcolor{blue!15}89.85 & \cellcolor{blue!15}151.09 ms & \cellcolor{olive!15}71.81 & \cellcolor{olive!15}90.25 & \cellcolor{olive!15}189.82 ms \\
        \bottomrule
    \end{tabular}
    \caption{Latency-constrained quantization on BISMO (edge and cloud accelerator) on ImageNet. Our framework can reduce the latency by \textbf{1.4$\times$} to \textbf{1.95$\times$} with negligible loss of accuracy compared with the fixed bitwidth (8 bits) quantization. }
    \label{tbl:bismo_latency}
\end{table*}
\begin{figure}[!t]
    \centering
    \includegraphics[width=1.05\linewidth]{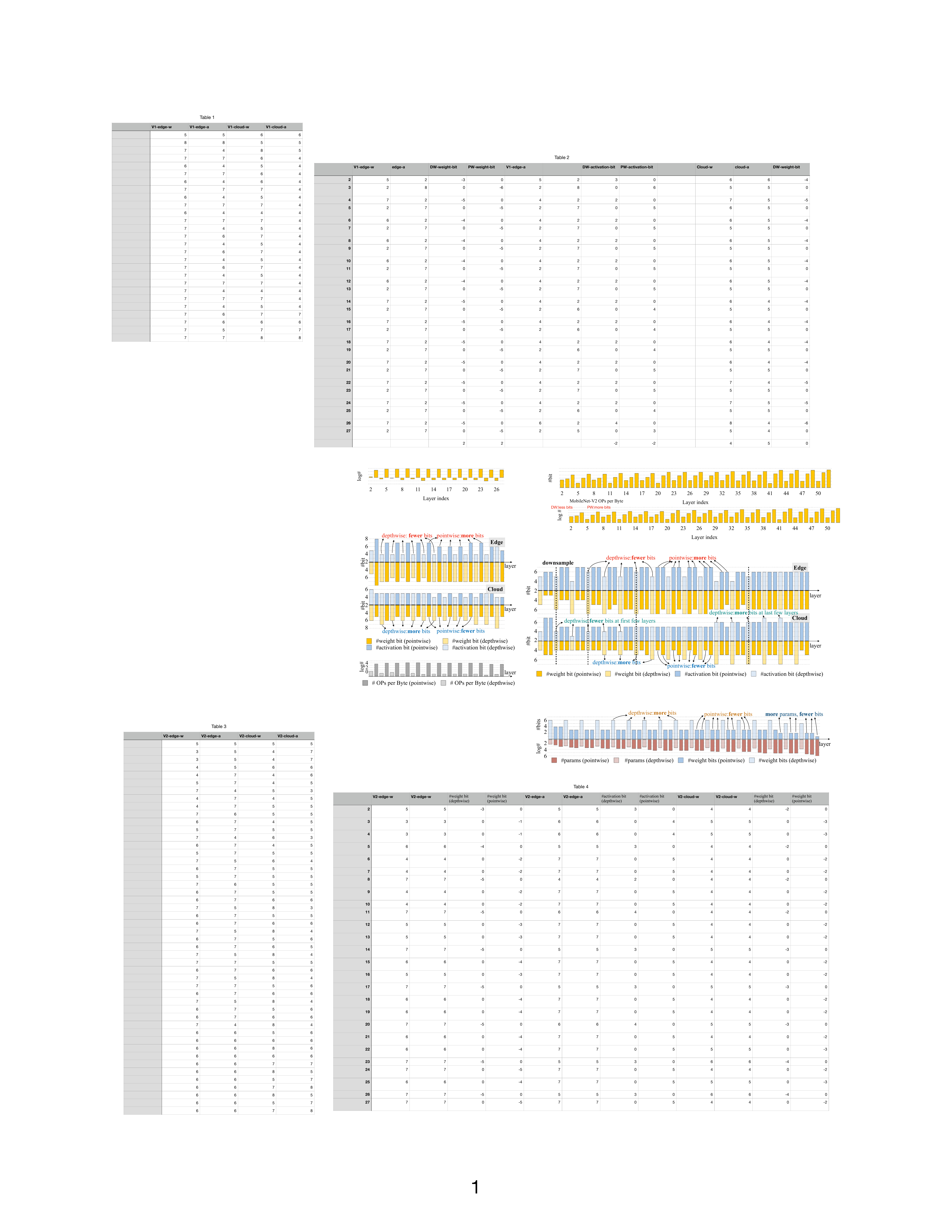}
    \caption{Quantization policy under latency constraints for MobileNet-V1 on BISMO (57.7 ms for the edge accelerator and 77.5 ms for the cloud accelerator). On edge accelerator, our agent allocates \emph{fewer} activation bits to the depthwise convolutions, which echos that the depthwise convolutions are memory bounded and the activations dominates the memory access. On cloud accelerator, our agent allocates \emph{more} bits to the depthwise convolutions and allocates \emph{fewer} bits to the pointwise convolutions, as cloud device has more memory bandwidth and higher parallelism, the network appears to be computation bounded.}
    \label{fig:bismo_v1}
\end{figure}
\begin{figure*}[!t]
    \centering
    \includegraphics[width=\linewidth]{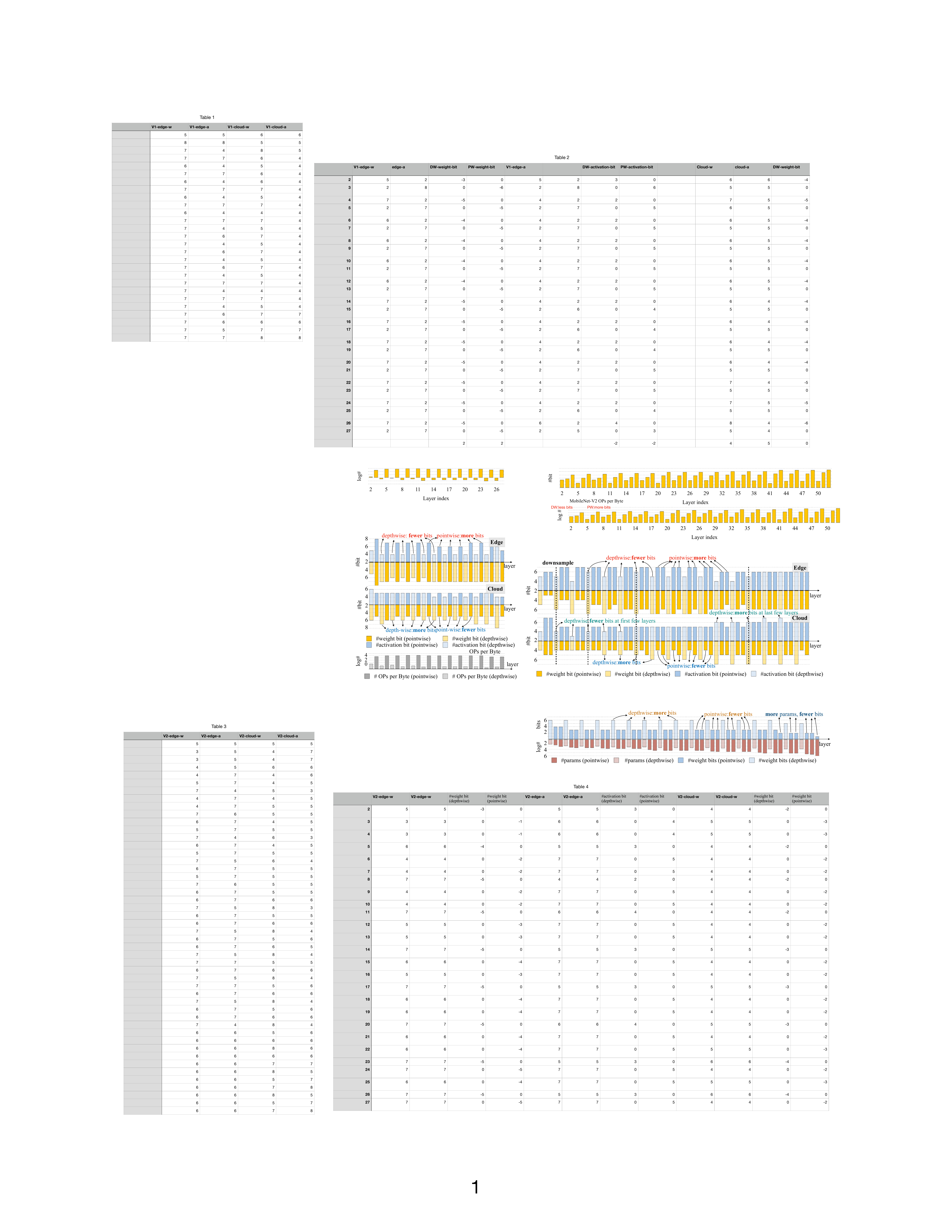}
    \caption{Quantization policy under latency constraints for MobileNet-V2 on BISMO (66.9 ms for the edge accelerator and 99.1 ms for the cloud accelerator). Similar to \fig{fig:bismo_v1}, depthwise layer is assigned with fewer bits on the edge accelerator, and pointwise layer is assigned with fewer bits on the cloud accelerator.}
    \label{fig:bismo_latency_v2}
\end{figure*}

\subsubsection{Quantization policy for BISMO Architecture}

Inferencing the neural networks on edge devices and cloud severs can be quite different, since the tasks on the cloud servers are more intensive and the edge devices are usually limited to low computation resources and memory bandwidth. We use Xilinx Zynq-7020 FPGA~\citep{zync7020} as our edge device and Xilinx VU9P~\citep{vu9p} as our cloud device. \tbl{tbl:edge_cloud} shows our experiment configurations on these two platforms along with their available resources.

As for comparison, we adopt the PACT~\citep{Choi:2018uw} as our baseline, which uses the same number of bits for all layers except for the first layer which extracts the low level features, they use 8 bits for both weights and activations as it has fewer parameters and is very sensitive to errors. We follow a similar setup as PACT: we quantize the weights and activations of the first and last layer to 8 bits and explore the bitwidth allocation policy for all the other layers.

Under the same latency, \modelshort consistently achieved better accuracy than the baseline on both the cloud and the edge (\tbl{tbl:bismo_latency}). With similar accuracy, \modelshort can reduce the latency by 1.4$\times$ to 1.95$\times$ compared with the baseline.

\myparagraph{Interpreting the quantization policy.}

Our agent gave quite different quantization policy for edge and cloud accelerators.
For the activations, the depthwise convolution layers are assigned much less bitwidth than the pointwise layers on the edge; while on the cloud device, the bitwidth of these two types of layers are similar to each other. For weights, the bitwidth of these types of layers are nearly the same on the edge; while on the cloud, the depthwise convolution layers are assigned much more bitwidth than the pointwise convolution layers.

\begin{table}[!t]
    \renewcommand*{\arraystretch}{1.2}
    \setlength{\tabcolsep}{5.5pt}
    \small\centering
    \begin{tabular}{lccccc}
        \toprule
        & Weights & Activations & Acc.-1 & Acc.-5 & Latency \\
        \midrule
        PACT & 4 bits & 4 bits & 62.44 & 84.19 & 7.86 ms \\
        Ours & \emph{flexible} & \emph{flexible} & \textbf{67.45} & \textbf{87.85} & 7.86 ms \\
        \midrule
        PACT & 6 bits & 4 bits & 67.51 & 87.84 & 11.10 ms \\
        Ours & \emph{flexible} & \emph{flexible} & \cellcolor{red!15}\textbf{70.40} & \cellcolor{red!15}\textbf{89.69} & \cellcolor{red!15}11.09 ms \\
        \midrule
        PACT & 6 bits & 6 bits & 70.46 & 89.59 & 19.99 ms \\
        Ours & \emph{flexible} & \emph{flexible} & \textbf{70.90} & \textbf{89.95} & 19.98 ms \\
        \midrule
        Original & 8 bits & 8 bits & \cellcolor{red!15}70.82 & \cellcolor{red!15}89.85 & \cellcolor{red!15}20.08 ms \\
        \bottomrule
    \end{tabular}
    \caption{Latency-constrained quantization on BitFusion (for MobileNet-V1 on ImageNet). Our \modelshort framework can reduce the latency by \textbf{2$\times$} with almost no loss of accuracy compared with the fixed bitwidth (8 bits) quantization.}
    \label{tbl:bitfusion_latency}
\end{table}
\begin{table}[!t]
    \renewcommand*{\arraystretch}{1.2}
    \setlength{\tabcolsep}{5.5pt}
    \small\centering
    \begin{tabular}{lccccc} 
        \toprule
        & Weights & Activations & Acc.-1 & Acc.-5 & Energy \\
        \midrule
        PACT & 4 bits & 4 bits & 62.44 & 84.19 & 13.47 mJ \\
        Ours & \emph{flexible} & \emph{flexible} & \textbf{64.78} & \textbf{85.85} & 13.69 mJ \\
        \midrule
        PACT & 6 bits & 4 bits & 67.51 & 87.84 & 16.57 mJ \\
        Ours & \emph{flexible} & \emph{flexible} & \cellcolor{red!15}\textbf{70.37} & \cellcolor{red!15}\textbf{89.40} & \cellcolor{red!15}16.30 mJ \\
        \midrule
        PACT & 6 bits & 6 bits & 70.46 & 89.59 & 26.80 mJ \\
        Ours & \emph{flexible} & \emph{flexible} & \textbf{70.90} & \textbf{89.73} & 26.67 mJ \\
        \midrule
        Original & 8 bits & 8 bits & \cellcolor{red!15}70.82 & \cellcolor{red!15}89.95 & \cellcolor{red!15}31.03 mJ \\
        \bottomrule
    \end{tabular}
    \caption{Energy-constrained quantization on BitFusion (for MobileNet-V1 on ImageNet). Our \modelshort framework reduces the power consumption by \textbf{2$\times$} with nearly no loss of accuracy compared with the fixed bitwidth quantization.}
    \label{tbl:bitfusion_energy}
\end{table}

We explain the difference of quantization policy between edge and cloud by the roofline model~\citep{williams2009roofline}. Many previous works use FLOPs or BitOPs as metrics to measure computation complexity. However, they are not able to directly reflect the latency, since there are many other factors influencing the hardware performance, such as memory access cost and degree of parallelism~\citep{Sandler:2018wy, liu2017learning}. Taking computation and memory access into account, the roofline model assumes that applications are either computation-bound or memory bandwidth-bound, if not fitting in on-chip caches, depending on their operation intensity. Operation intensity is measured as operations (MACs in neural networks) per DRAM byte accessed. A lower operation intensity indicates that the model suffers more from the memory access.

\begin{figure}[h]
    \centering
    \includegraphics[width=\linewidth]{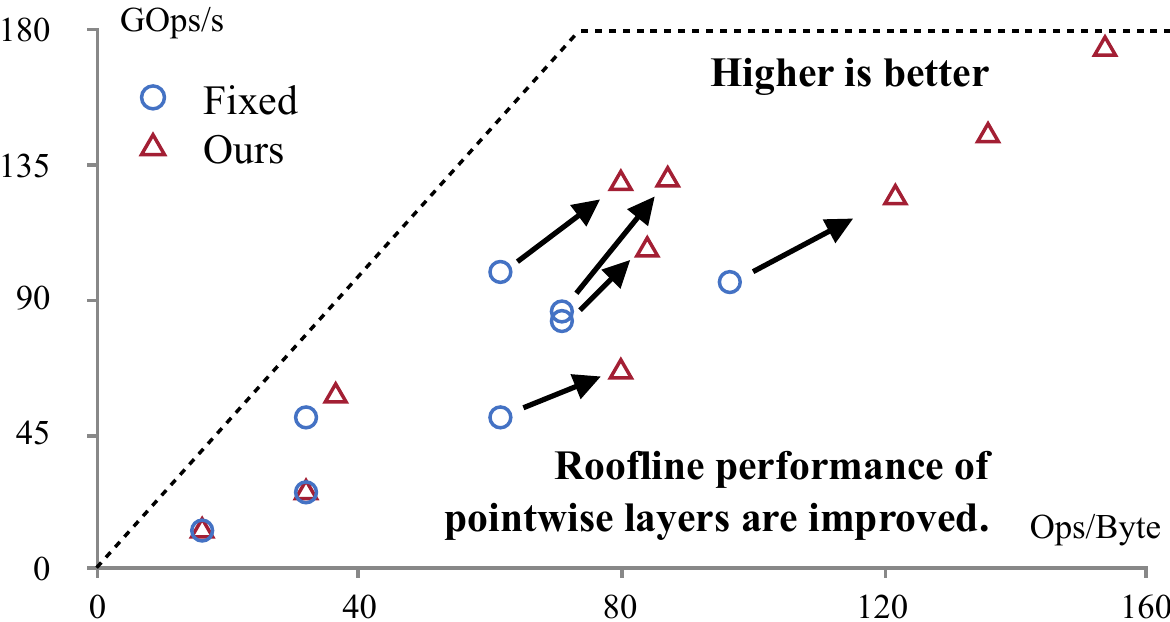}
    \caption{Roofline model of pointwise layers in MobileNet-V1 (fixed-bitwidth in blue and mixed-precision in red). Our mixed-precision framework improves the roofline performance by a large margin.}
    \label{fig:bismo_roofline}
\end{figure}

The bottom of \fig{fig:bismo_v1} shows the operation intensities (operations per byte) of convolution layers in the MobileNet-V1. Depthwise convolution is a memory bounded operation, and the pointwise convolution is a computation bounded operation. Our experiments show that when running MobileNet-V1 on the edge devices with small batch size, its latency is dominated by the depthwise convolution layers. Since the feature maps take a major proportion in the memory of depthwise convolution layers, our agent gives the activations fewer bits. In contrast, when running MobileNet-V1 on the cloud with large batch size, both two types of layers have nearly the equal influence on the speed. Therefore, our agent tries to reduce the bitwidth of both activation and weights. However, since the weights of the depthwise convolution layers takes a small proportion of the memory, our agent increases their bitwidth to preserve the network accuracy at low memory overhead. \fig{fig:bismo_roofline} shows the roofline model before and after HAQ. HAQ gives more reasonable policy to allocate the bits for each layer and pushes all the points to the upper right corner that is more efficient.

On edge accelerator, which has much less memory bandwidth, our RL agent allocates \emph{fewer} activation bits to the depthwise convolutions since the activations dominates the memory access. On cloud accelerator, which has more memory bandwidth, our agent allocates \emph{more} bits to the depthwise convolutions and allocates \emph{fewer} bits to the pointwise convolutions to prevent it from being computation bounded.

A similar phenomenon can be observed in \fig{fig:bismo_latency_v2} for quantizing MobileNet-V2. Moreover, since the activation size in the deeper layers gets smaller, they get assigned more bits. Another interesting phenomenon we discover in \fig{fig:bismo_latency_v2} is that the downsample layer gets assigned more activation bits than the adjacent layer. This is because downsampled layers are more prone to lose information, so our agent learns to assign more bits to the activations to compensate.

\subsubsection{Quantization policy for BitFusion Architecture}

In order to demonstrate the effectiveness of our framework on different hardware architectures, we further compare our framework with PACT~\citep{Choi:2018uw} under the latency constraints on the BitFusion~\citep{sharma2018bit} architecture. As demonstrated in \tbl{tbl:bitfusion_latency}, our framework performs much better than the hand-craft policy with the same latency. Also, it can achieve almost no degradation of accuracy with only half of the latency used by the original MobileNet-V1 model (from \textbf{20.08} to \textbf{11.09} ms). Therefore, our framework is indeed very flexible and can be applied to different hardware platforms.

\subsection{Energy-Constrained Quantization}

We then evaluate our framework under the energy constraints on the BitFusion~\citep{sharma2018bit} architecture. Similar to the latency-constrained experiments, we compare our framework with PACT~\citep{Choi:2018uw} which uses fixed number of bits for both weights and activations. From \tbl{tbl:bitfusion_energy}, we can clearly see that our framework outperforms the rule-based baseline: it achieves much better performance while consuming similar amount of energy. In particular, our framework is able to achieve almost no loss of accuracy with nearly half of the energy consumption of the original MobileNet-V1 model (from \textbf{31.03} to \textbf{16.57} mJ), which suggests that flexible bitwidths can indeed help reduce the energy consumption.

\subsection{Model Size-Constrained Quantization}

\begin{table*}[!t]
    \renewcommand*{\arraystretch}{1.2}
    \setlength{\tabcolsep}{7pt}
    \small\centering
    \begin{tabular}{lcccccccccccc} 
        \toprule
        & & \multicolumn{3}{c}{MobileNet-V1} & \multicolumn{3}{c}{MobileNet-V2} & \multicolumn{3}{c}{ResNet-50} \\
        \cmidrule(lr){3-5}\cmidrule(lr){6-8}\cmidrule(lr){9-11}
        & Weights & Acc.-1 & Acc.-5 & Model Size & Acc.-1 & Acc.-5 & Model Size & Acc.-1 & Acc.-5 & Model Size \\
        \midrule
        \cite{Han:2016uf} & 2 bits & 37.62 & 64.31 & 1.09 MB & 58.07 & 81.24 & 0.96 MB & 68.95 & 88.68 & 6.32 MB \\
        Ours & \emph{flexible} & \textbf{57.14} & \textbf{81.87} & 1.09 MB & \textbf{66.75} & \textbf{87.32} & 0.95 MB & \textbf{70.63} & \textbf{89.93} & 6.30 MB \\
        \midrule
        \cite{Han:2016uf} & 3 bits & 65.93 & 86.85 & 1.60 MB & 68.00 & 87.96 & 1.38 MB & 75.10 & 92.33 & 9.36 MB \\
        Ours & \emph{flexible} & \textbf{67.66} & \textbf{88.21} & 1.58 MB & \textbf{70.90} & \textbf{89.76} & 1.38 MB & \textbf{75.30} & \textbf{92.45} & 9.22 MB \\
        \midrule
        \cite{Han:2016uf} & 4 bits  & 71.14 & 89.84 & 2.10 MB & 71.24 & 89.93 & 1.79 MB & \textbf{76.15} & 92.88 & 12.40 MB \\
        Ours & \emph{flexible} & \textbf{71.74} & \textbf{90.36} & 2.07 MB & \textbf{71.47} & \textbf{90.23} & 1.79 MB & 76.14 & \textbf{92.89} & 12.14 MB \\
        \midrule
        Original & 32 bits & 70.90 & 89.90 & 16.14 MB & 71.87 & 90.32 & 13.37 MB & 76.15 & 92.86 & 97.49 MB \\
        \bottomrule
    \end{tabular}
    \caption{Model size-constrained quantization on ImageNet. Compared with Deep Compression~\citep{Han:2017td}, our framework achieves higher accuracy under similar model size (especially under high compression ratio).}
    \label{tbl:model_size}
\end{table*}
\begin{figure*}[!t]
    \centering
    \includegraphics[width=\linewidth]{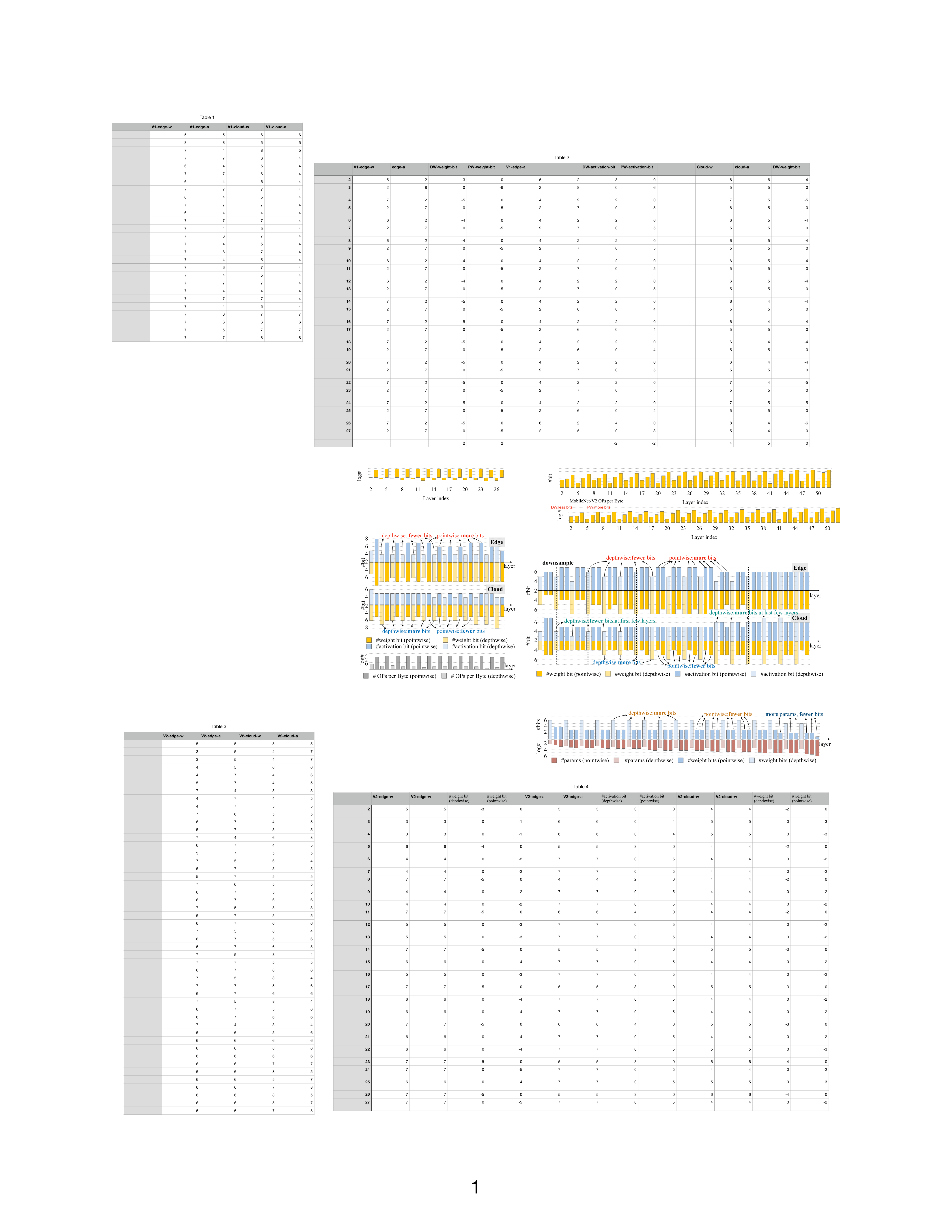}
    \caption{Quantization policy under model size constraints for MobileNet-V2. Our RL agent allocates \emph{more} bits to the depthwise convolutions, since depthwise convolutions have \emph{fewer} number of parameters.}
    \label{fig:model_size}
\end{figure*}

We further evaluate our \modelshort framework under the model size constraints. Following \cite{Han:2016uf}, we employ the $k$-means algorithm to quantize the values into $k$ centroids instead of using the linear quantization for compression.

We compare our framework with Deep Compression~\citep{Han:2016uf} on MobileNets and ResNet-50. From \tbl{tbl:model_size}, we can see that our framework performs much better than Deep Compression: it achieves higher accuracy with the same model size. For MobileNets which are already very compactly designed, our framework can preserve the performance to some extent; while Deep Compression significantly degrades the performance especially when the model size is very small. For instance, when Deep Compression quantizes the weights of MobileNet-V1 to 2 bits, the accuracy drops significantly from 70.90 to \textbf{37.62}; while our framework can still achieve \textbf{57.14} of accuracy with the same model size, which is because our framework makes full use of the flexible bitwidths.

\begin{figure}[htb]
    \centering
    \includegraphics[width=1.0\linewidth]{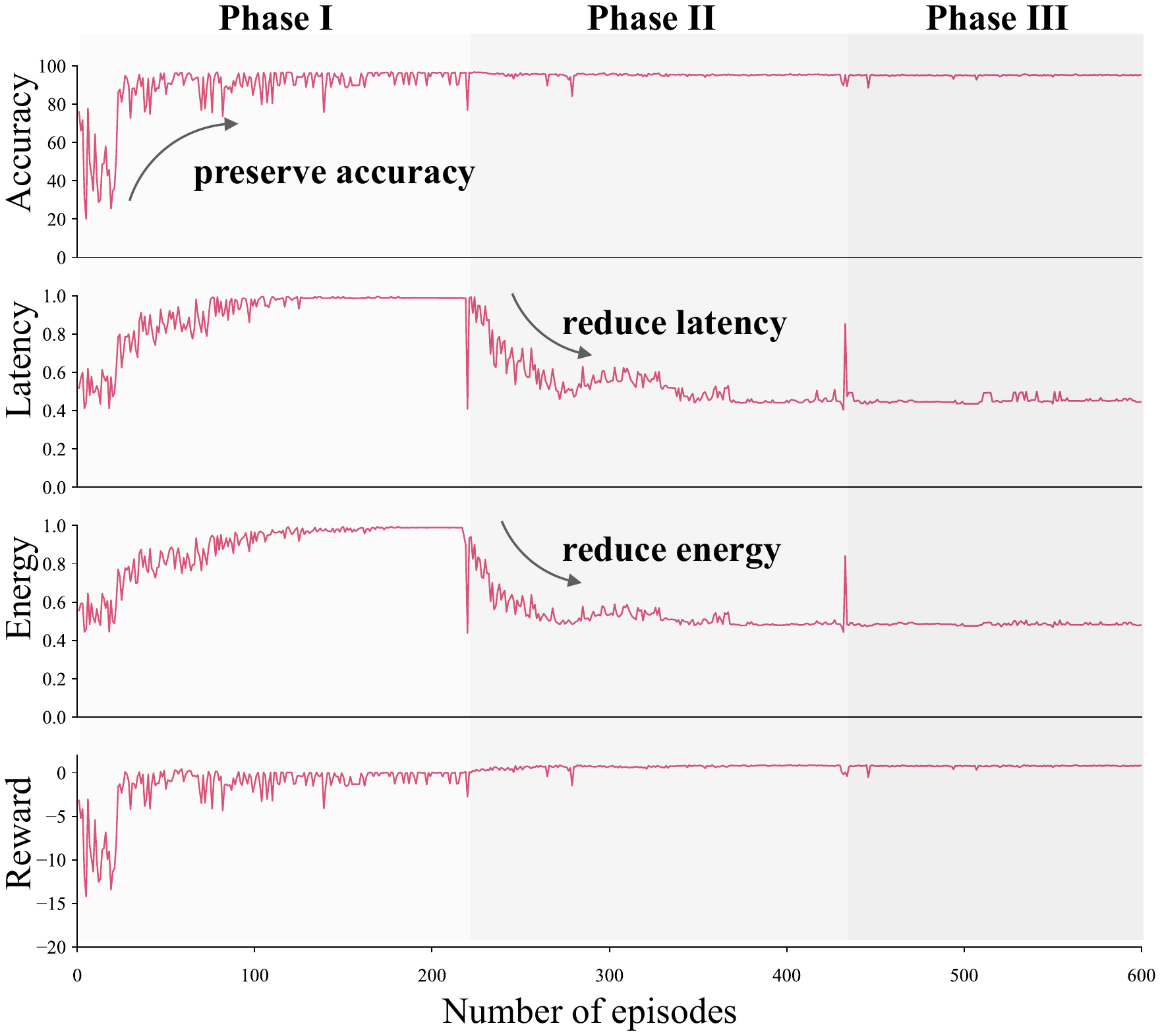}
    \caption{Exploration curves of accuracy-guaranteed quantization for MobileNet-V1. Our RL agent first tries to preserve the accuracy while completely ignoring the latency and the energy consumption; after the accuracy begins to be more stable, it starts to aggressively reduce the latency and the energy; it finally converges to the best policy it has found.}
    \label{fig:accuracy}
\end{figure}

\myparagraph{Discussions.}

In \fig{fig:model_size}, we visualize the bitwidth allocation strategy for MobileNet-V2. From this figure, we can observe that our framework assigns \emph{more} bitwidths to the weights in depthwise convolution layers than pointwise convolution layers. Intuitively, this is because the number of parameters in the former is much smaller than the latter. Comparing \fig{fig:bismo_latency_v2} and \fig{fig:model_size}, the policies are drastically different under different optimization objectives (\textbf{fewer} bitwiths for depthwise convolutions under \emph{latency} optimization, \textbf{more} bitwidths for depthwise convolutions under \emph{model size} optimization). Our framework succeeds in learning to adjust its bitwidth policy under different constraints.

\subsection{Accuracy-Guaranteed Quantization}

Apart from the resource-constrained experiments, we also evaluate our framework under the accuracy-guaranteed scenario, that is to say, we aim to minimize the resource (\ie, latency and energy) we use while preserving the accuracy.

Instead of using the resource-constrained action space in \sect{sect:approach:action_space}, we define a new reward function $R$ that takes both the resource and the accuracy into consideration:
\begin{equation}
    R = R_\text{latency} + R_\text{energy} + 
    R_\text{accuracy}.
\end{equation}
Here, the reward functions $R_\text{*}$ are defined to encourage each term to be as good as possible:
\begin{equation}
\begin{aligned}
    R_\text{latency} &= \lambda_\text{latency} \times (\text{latency}_\text{quant} - \text{latency}_\text{origin}), \\
    R_\text{energy} &= \lambda_\text{energy} \times (\text{energy}_\text{quant} - \text{energy}_\text{origin}), \\
    R_\text{accuracy} &= \lambda_\text{accuracy} \times (\text{accuracy}_\text{quant} - \text{accuracy}_\text{origin}),
\end{aligned}
\end{equation}
where $\lambda_\text{*}$ are scaling factors that encourage the RL agent to trade off between the computation resource and the accuracy. We set $\lambda_\text{latency}$ and $\lambda_\text{energy}$ to $1$, and $\lambda_\text{accuracy}$ to $20$ in our experiments to ensure that our RL agent will prioritize the accuracy over the computation resource.

We choose to perform our experiments on a ten-category subset of ImageNet as it is very challenging to preserve the accuracy while reducing the computation resource. In \fig{fig:accuracy}, we illustrate the exploration curves of our RL agents, and we can observe that the exploration process can be divided into three phases. In the first phase, our RL agent puts its focus on the accuracy: it tries to preserve the accuracy while completely ignoring the latency and the energy consumption. In the second phase, the accuracy begins to be more stable, and our RL agent starts to aggressively reduce the latency and the energy. In the third phase, our RL agent converges to the best policy it has found. We conjecture that this interesting behavior is because that the scaling factor $\lambda_\text{accuracy}$ is much larger than the other two, which encourages our agent to first optimize the value of accuracy, and after its value has been stabilized, our agent then attempts to reduce the value of latency and energy to further optimize the reward value (see the reward curve in \fig{fig:accuracy}).

\subsection{Integration with Architecture Search and Pruning}

\begin{figure}[!t]
    \centering
    \includegraphics[width=0.93\linewidth]{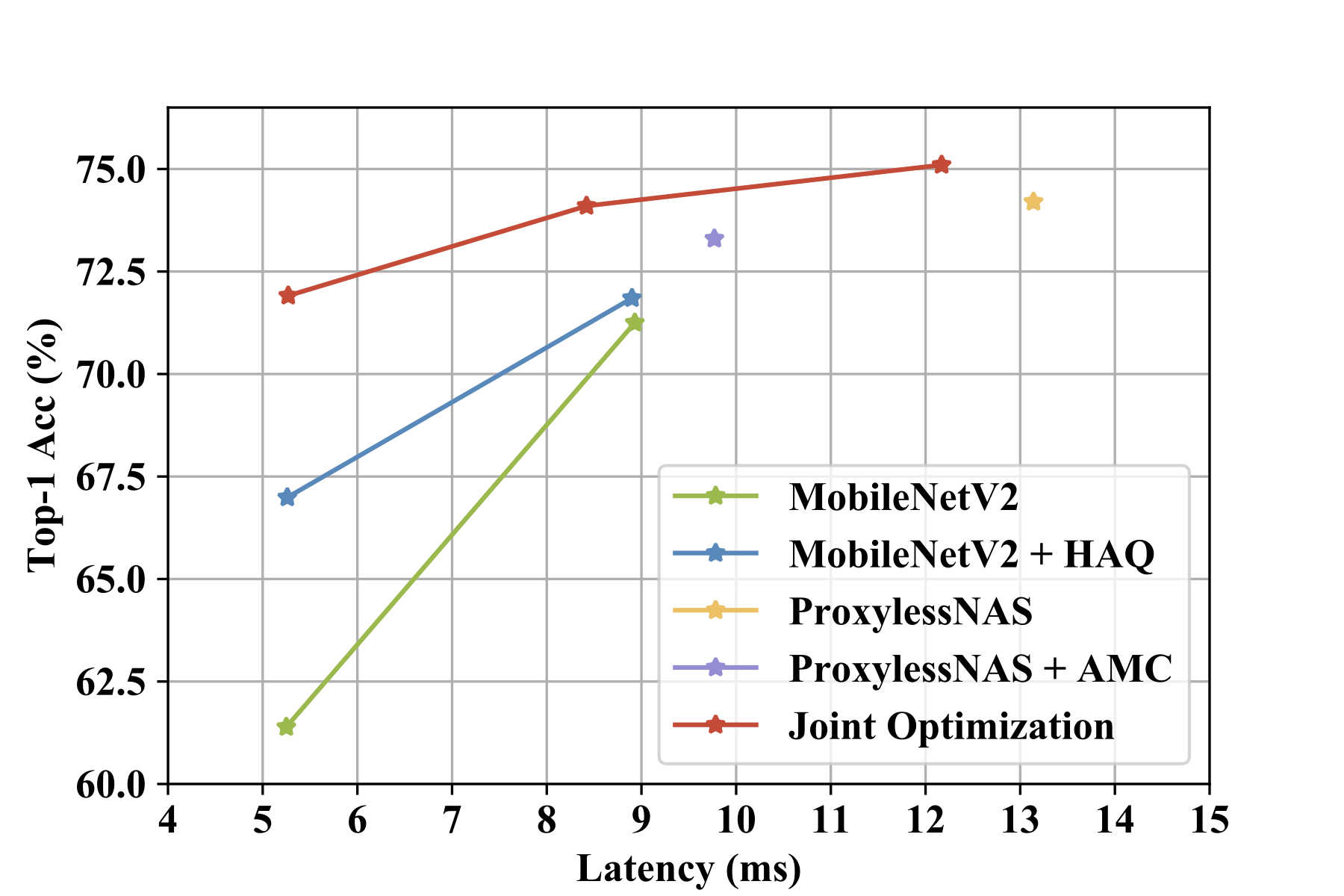}
    \caption{Integrating NAS and AMC with HAQ together further improves the accuracy-latency trade-off by a significant margin.}
    \label{fig:bitfusion_latency_pipeline}
\end{figure}

We integrate the neural architecture search~\citep{Cai:2019ui} and automated channel pruning~\citep{He:2018vj} with HAQ to demonstrate that our method is orthogonal to other AutoML methods. In \fig{fig:bitfusion_latency_pipeline}, we observe significant improvements over the baselines including ProxylessNAS (with 8-bit quantization), ProxylessNAS + AMC (with 8-bit quantization), MobileNetV2 (with 4-bit / 6-bit quantization), and MobileNetV2 + HAQ (with mixed-precision quantization).

\section{Analysis}

In this section, we first compare with sample efficiency of different optimization methods; then, we show the generalization and transfer learning ability of our framework; we finally interpret the quantization policy given by HAQ.

\subsection{Optimization Methods}

\begin{table}[!t]
    \renewcommand*{\arraystretch}{1.2}
    \setlength{\tabcolsep}{8pt}
    \small\centering
    \begin{tabular}{lcccc}
        \toprule
        & Search Time & Acc.-1 & Acc.-5 & Latency \\
        \midrule
        ES & \textbf{17 hours} & 65.73 & 86.81 & 45.45 ms \\
        BO & 74 hours & 66.28 & 87.22 & 45.47 ms \\
        RL (Ours) & \textbf{17 hours} & \textbf{67.40} & \textbf{87.90} & 45.51 ms \\
        \midrule
        ES & \textbf{17 hours} & 69.11 & 88.80 & 57.73 ms \\
        BO & 74 hours & 70.40 & 89.56 & 57.68 ms \\
        RL (Ours) & \textbf{17 hours} & \textbf{70.58} & \textbf{89.77} & 57.70 ms \\
        \bottomrule
    \end{tabular}
    \caption{Comparison between different optimization methods. RL outperforms EA and BO and is \textbf{4$\times$} faster than BO in terms of the total search time.}
    \label{tbl:bismo_optimizer}
\end{table}
\begin{figure*}[h]
    \centering
    \begin{subfigure}[b]{0.99\linewidth}
         \includegraphics[width=\linewidth]{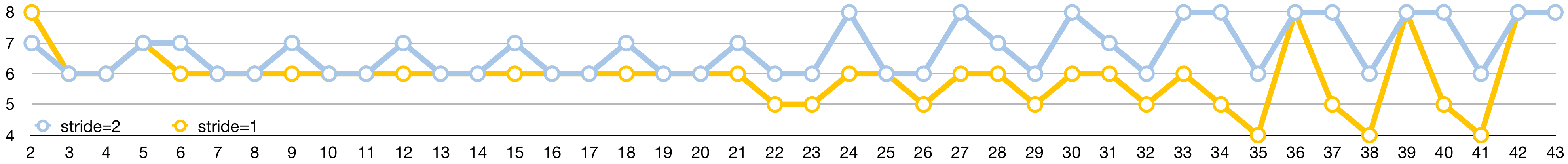}
         \caption{The agent allocates \textbf{more activation bits} to \textbf{stride=2} layers, which compensates for the information loss from downsampling.}
    \end{subfigure}
    \begin{subfigure}[b]{0.99\linewidth}
         \includegraphics[width=\linewidth]{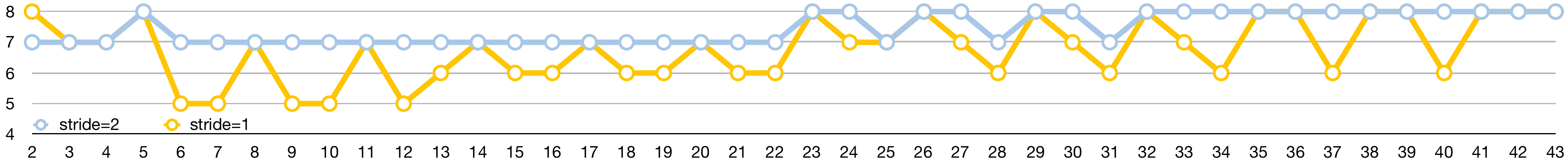}
         \caption{The agent allocates \textbf{more weight bits} to \textbf{stride=2} layers, which compensates for the information loss from downsampling.}
    \end{subfigure}
    \begin{subfigure}[b]{0.99\linewidth}
         \includegraphics[width=\linewidth]{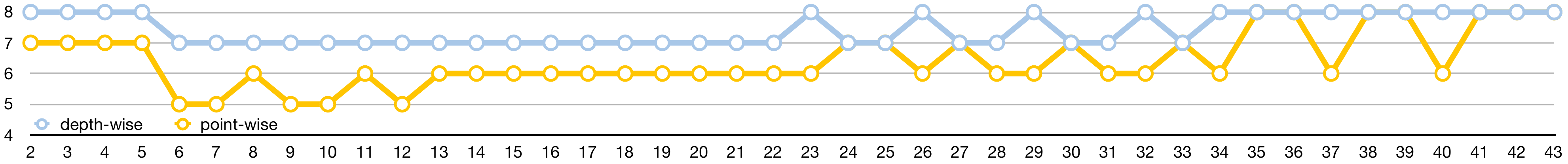}
         \caption{The agent allocates \textbf{more weight bits} to \textbf{depthwise} layers than \textbf{pointwise} layers, because depthwise layers have fewer weights.}
    \end{subfigure}
    \caption{We change only one dimension of the state vector, and run the actor network again to observe how the action changes across different layers. }
    \label{fig:difference}
\end{figure*}
\begin{figure}[b]
    \centering
    \includegraphics[width=\linewidth]{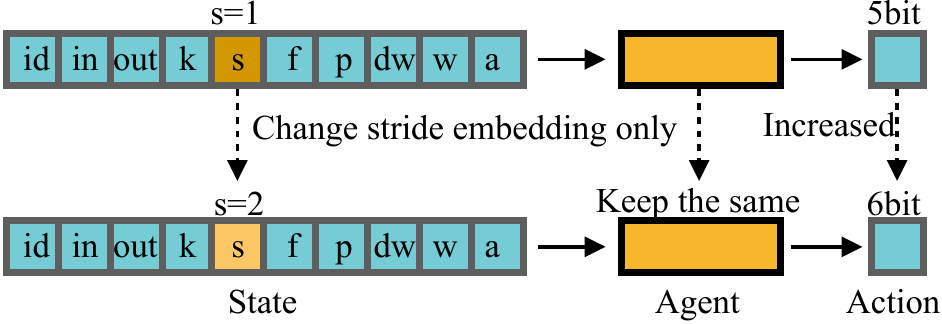}
    \caption{Policy interpretation. From the model, we first select several layers; then, we only change (flip) one dimension in the state vector; finally, we run our RL agent's actor network again to see how that particular factor affects its dicision.}
    \label{fig:interpretation}
\end{figure}

We leverage the reinforcement learning (RL) as our optimization method. In addition, we also compared with other optimizers including Bayesian optimization (BO) and evolutionary algorithm (EA). Similar to the configurations of RL, we model the outputs of BO and EA as the number of bits of different layers and the objectives of BO and EA as maximizing the validation accuracy of the quantized model.

As a fair comparison, we executed in total 600 runs (samples) for each optimization method. The performance comparison is in \tbl{tbl:bismo_optimizer}. All experiments are conducted on the BISMO hardware with MobileNet-V1. We observe that RL performs much better than EA and BO and is \textbf{4$\times$} faster than BO in terms of the total search time. This, we believe, is because BO and EA do not make use of the state encoding, which might lead to a worse sample efficiency.

\subsection{Generalization and Transfer Learning}

\begin{table}[!t]
    \renewcommand*{\arraystretch}{1.2}
    \setlength{\tabcolsep}{5pt}
    \small\centering
    \begin{tabular}{lcccc}
        \toprule
        & Bitwidth & Acc.-1 & Acc.-5 & Latency \\
        \midrule
        PACT & 4 bits & 61.39 & 83.72 & 52.15 ms \\
        Ours (search for V2) & \emph{flexible} & 66.99 & 87.33 & 52.12 ms \\
        Ours (transfer from V1) & \emph{flexible} & 65.80 & 86.60 & 52.06 ms \\
        \midrule
        PACT & 5 bits & 68.84 & 88.58 & 66.94 ms \\
        Ours (search for V2) & \emph{flexible} & 70.90 & 89.91 & 66.92 ms \\
        Ours (transfer from V1) & \emph{flexible} & 69.90 & 89.24 & 66.93 ms \\
        \bottomrule
    \end{tabular}
    \caption{Comparisons between our agent's transfer results (from MobileNet-V1 to MobileNet-V2), its direct search results on MobileNet-V2, and the fixed-bitwidth baseline (PACT). Our RL agent is able to generalize well to different network architectures: our quantization policy transferred from V1 to V2 performs better than the fixed-bitwidth baseline and is only slighly worse than the quantization policy directly searched for V2.}
    \label{tbl:bismo_transfer}
\end{table}

Another merit of the reinforcement learning is that its agent is able to generalize to different environments (\ie, network architectures). In order to evaluate the transfer ability of our framework, we first train our agent for MobileNet-V1 under the latency constraint, and we then directly evaluate our agent on MobileNet-V2 by feeding its network architecture information in. In \tbl{tbl:bismo_transfer}, we compare our agent's transfer results (from V1 to V2) with its direct search results (for V2) and the fixed-bitwidth baseline (PACT). Our quantization policy transferred from V1 to V2 still performs better than the fixed-bitwidth baseline and is only slightly worse than the quantization policy directly searched for V2. This experiment validates that our RL agent generalize well to different network architectures.

\subsection{Importantance of Hardware-Awareness}

\begin{table}[!t]
    \renewcommand*{\arraystretch}{1.2}
    \setlength{\tabcolsep}{5.5pt}
    \small\centering
    \begin{tabular}{lccccc} 
        \toprule
        Constraint & Bitwidth & BitOPs & latency & Acc.-1 & Acc.-5 \\
        \midrule
        BitOPs & \emph{flexible} & 8.17 G & 85.06 ms & 70.29 & 89.52 \\
        Latency & \emph{flexible} & \textbf{8.01 G} & \textbf{66.92 ms} & \textbf{70.90} & \textbf{89.91} \\
        \midrule
        BitOPs & \emph{flexible} & 11.36 G & 97.99 ms & 71.41 & 90.12 \\
        Latency & \emph{flexible} & \textbf{11.17 G} & \textbf{82.34 ms} & \textbf{71.89} & \textbf{90.36} \\
        \bottomrule
    \end{tabular}
    \caption{BitOPs-constrained quantization on BISMO (MobileNet-V2 on ImageNet).}
    \label{tbl:bitops}
\end{table}

To evaluate the necessity of involving the hardware in the loop, we replace the BISMO accelerator with theoretical BitOPs analysis, which calculates the latency by $\text{FLOPs}/\text{s} \times Bit_\text{weight} \times Bit_\text{activation}$ for each layer. The results are listed in Table~\ref{tbl:bitops}, which shows that under similar BitOPs constrains, BitOPs constrained experiments get worse latency than the experiments with hardware-in-the-loop, and BitOPs constrained experiments could achieve better accuracy when the constrain is tight. The reason is that BitOPs and hardware latency are not linearly correlated, so the similar BitOPs may correspond to totally different latency if the layer is memory bottlenecked. The agent chooses to give more bits to depthwise layers which have less FLOPs, but more memory access (which means more latency in edge hardware).

\subsection{Performance on Large Model}

\begin{table}[!t]
    \renewcommand*{\arraystretch}{1.2}
    \setlength{\tabcolsep}{5.5pt}
    \small\centering
    \begin{tabular}{lccccc}
        \toprule
        & Weights & Activations & Acc.-1 & Acc.-5 & Latency \\
        \midrule
        PACT & 2 bits & 4 bits & 74.06 & 91.78 & 80.03 ms \\
        Ours & \emph{flexible} & \emph{flexible} & \textbf{74.42} & \textbf{91.92} & 80.38 ms \\
        \midrule
        PACT & 3 bits & 3 bits & 74.73 & 92.11 & 85.49 ms \\
        Ours & \emph{flexible} & \emph{flexible} & \textbf{74.98} & \textbf{92.37} & 84.97 ms \\
        \midrule
        PACT & 4 bits & 4 bits & 76.17 & 93.03 & 128.55 ms \\
        Ours & \emph{flexible} & \emph{flexible} & \textbf{76.22} & \textbf{93.15} & 129.55 ms \\
        \midrule
        Original & 8 bits & 8 bits & 76.64 & 93.26 & 446.96 ms \\
        \bottomrule
    \end{tabular}
    \caption{Latency-constrained quantization on BISMO (for ResNet-50 on ImageNet).}
    \label{tbl:bismo_resnet50}
\end{table}

We evaluate our framework on a larger ResNet-50 model with the same search scheme and finetune policy as in Table~\ref{tbl:bismo_latency}, and the hardware platform is BISMO edge hardware simulator. Table~\ref{tbl:bismo_resnet50} shows that the improvement of HAQ is not remarkable on the large model like ResNet50. The reason is that ResNet-50 is highly redundant, even PACT can already quantize it to 4 bits without much accuracy loss. Therefore, there is not much room for HAQ to further improve it.

\subsection{Policy Interpretation}

In \sect{sect:exp}, we provided intuitive explanations of our agent's policies. In this section, we quantitatively interpret our agent's quantization policy. As illustrated in \fig{fig:interpretation}, we first select several layers from MobileNet-V2; then for each layer, we change (flip) only one dimension in the state vector (in the example, changing the convolution stride from 1 to 2); finally, we run feedforward on our RL agent's actor network again to see how that particular factor (\ie, depthwise, downsample) affects its decision.

From \fig{fig:difference}, we can clearly observe that some factors will affect the actions. For instance, if we only change the stride embedding of each layer in MobileNet-V2, we could observe that our agent will allocate more activation bits to downsample layers (stride=2), and this phenomenon is more obviously at deep layers. As for weight bits, our agent also allocate more bits for downsample layers. Moreover, if we only change the depthwise/pointwise embedding, we could observe that pointwise layers will be allocated fewer weight bits, the reason may lay on pointwise layer are more computation bounded, fewer weight bits will obviously reduce the computation complexity.

\section{Conclusion}

In this paper, we propose an automated framework for quantization, \model (\modelshort), which does not require any domain experts and rule-based heuristics. We provide a learning-based method that can search the quantization policy with hardware feedback. Compared with indirect proxy signals, our framework can offer a specialized quantization solution for different hardware platforms. Extensive experiments demonstrate that our framework performs better than conventional rule-based approaches for multiple objectives: latency, energy and model size. Our framework reveals that the optimal policies on different hardware architectures are drastically different, and we interpreted the implication of those policies. We believe the insights will inspire the future software and hardware co-design for efficient deployment of deep neural networks. 
\section*{Acknowledgements}

We thank NSF Career Award \#1943349, MIT-IBM Watson AI Lab, Samsung, SONY, Xilinx, TI and AWS for supporting this research.

\bibliographystyle{spbasic}
\bibliography{reference}

\begin{thebibliography}{35}
\providecommand{\natexlab}[1]{#1}
\providecommand{\url}[1]{{#1}}
\providecommand{\urlprefix}{URL }
\expandafter\ifx\csname urlstyle\endcsname\relax
  \providecommand{\doi}[1]{DOI~\discretionary{}{}{}#1}\else
  \providecommand{\doi}{DOI~\discretionary{}{}{}\begingroup
  \urlstyle{rm}\Url}\fi
\providecommand{\eprint}[2][]{\url{#2}}

\bibitem[{Apple(2018)}]{apple}
Apple (2018) Apple describes 7nm {A12} bionic chips.
  \urlprefix\url{http://www.eenewsanalog.com/news/apple-describes-7nm-a12-bionic-chip/page/0/1}

\bibitem[{Cai et~al(2018)Cai, Yang, Zhang, Han, and Yu}]{Cai:2018wb}
Cai H, Yang J, Zhang W, Han S, Yu Y (2018) {Path-Level Network Transformation
  for Efficient Architecture Search.} In: ICML

\bibitem[{Cai et~al(2019)Cai, Zhu, and Han}]{Cai:2019ui}
Cai H, Zhu L, Han S (2019) {ProxylessNAS: Direct Neural Architecture Search on
  Target Task and Hardware}. In: ICLR

\bibitem[{Choi et~al(2018)Choi, Wang, Venkataramani, Chuang, Srinivasan, and
  Gopalakrishnan}]{Choi:2018uw}
Choi J, Wang Z, Venkataramani S, Chuang PIJ, Srinivasan V, Gopalakrishnan K
  (2018) {PACT: Parameterized Clipping Activation for Quantized Neural
  Networks}. arXiv

\bibitem[{Chollet(2017)}]{Chollet:2017vb}
Chollet F (2017) {Xception - Deep Learning with Depthwise Separable
  Convolutions.} In: CVPR

\bibitem[{Courbariaux et~al(2016)Courbariaux, Hubara, Soudry, El-Yaniv, and
  Bengio}]{Courbariaux:2016tm}
Courbariaux M, Hubara I, Soudry D, El-Yaniv R, Bengio Y (2016) {Binarized
  Neural Networks: Training Deep Neural Networks with Weights and Activations
  Constrained to +1 or -1}. arXiv

\bibitem[{Deng et~al(2009)Deng, Dong, Socher, Li, Li, and Li}]{Deng:2009td}
Deng J, Dong W, Socher R, Li LJ, Li K, Li FF (2009) {ImageNet - A large-scale
  hierarchical image database.} In: CVPR

\bibitem[{Han(2017)}]{Han:2017td}
Han S (2017) {Efficient Methods and Hardware for Deep Learning}. PhD thesis

\bibitem[{Han et~al(2016)Han, Mao, and Dally}]{Han:2016uf}
Han S, Mao H, Dally W (2016) {Deep Compression: Compressing Deep Neural
  Networks with Pruning, Trained Quantization and Huffman Coding}. In: ICLR

\bibitem[{He et~al(2016)He, Zhang, Ren, and Sun}]{He:2015tt}
He K, Zhang X, Ren S, Sun J (2016) {Deep Residual Learning for Image
  Recognition}. In: CVPR

\bibitem[{He et~al(2017)He, Zhang, and Sun}]{he2017channel}
He Y, Zhang X, Sun J (2017) Channel pruning for accelerating very deep neural
  networks. In: ICCV

\bibitem[{He et~al(2018)He, Lin, Liu, Wang, Li, and Han}]{He:2018vj}
He Y, Lin J, Liu Z, Wang H, Li LJ, Han S (2018) {AMC: AutoML for Model
  Compression and Acceleration on Mobile Devices}. In: ECCV

\bibitem[{Howard et~al(2017)Howard, Zhu, Chen, Kalenichenko, Wang, Weyand,
  Andreetto, and Adam}]{Howard:2017wz}
Howard AG, Zhu M, Chen B, Kalenichenko D, Wang W, Weyand T, Andreetto M, Adam H
  (2017) {MobileNets: Efficient Convolutional Neural Networks for Mobile Vision
  Applications}. arXiv

\bibitem[{Imagination(2018)}]{imagination}
Imagination (2018) Powervr neural network accelerator.
  \urlprefix\url{https://www.imgtec.com/vision-ai/powervr-series2nx/powervr-ax2145-nna/}

\bibitem[{Jacob et~al(2018)Jacob, Kligys, Chen, Zhu, Tang, Howard, Adam, and
  Kalenichenko}]{Jacob:2018ur}
Jacob B, Kligys S, Chen B, Zhu M, Tang M, Howard AG, Adam H, Kalenichenko D
  (2018) {Quantization and Training of Neural Networks for Efficient
  Integer-Arithmetic-Only Inference.} In: CVPR

\bibitem[{Kingma and Ba(2015)}]{Kingma:2015us}
Kingma D, Ba J (2015) {Adam - A Method for Stochastic Optimization.} In: ICLR

\bibitem[{Krishnamoorthi(2018)}]{Krishnamoorthi:2018wr}
Krishnamoorthi R (2018) {Quantizing deep convolutional networks for efficient
  inference - A whitepaper.} arXiv

\bibitem[{Lillicrap et~al(2016)Lillicrap, Hunt, Pritzel, Heess, Erez, Tassa,
  Silver, and Wierstra}]{Lillicrap:2016ww}
Lillicrap T, Hunt JJ, Pritzel A, Heess N, Erez T, Tassa Y, Silver D, Wierstra D
  (2016) {Continuous control with deep reinforcement learning.} In: ICLR

\bibitem[{Liu et~al(2018)Liu, Zoph, Neumann, Shlens, Hua, Li, Fei-Fei, Yuille,
  Huang, and Murphy}]{Liu:2018tr}
Liu C, Zoph B, Neumann M, Shlens J, Hua W, Li LJ, Fei-Fei L, Yuille A, Huang J,
  Murphy K (2018) {Progressive Neural Architecture Search}. In: ECCV

\bibitem[{Liu et~al(2017)Liu, Li, Shen, Huang, Yan, and
  Zhang}]{liu2017learning}
Liu Z, Li J, Shen Z, Huang G, Yan S, Zhang C (2017) Learning efficient
  convolutional networks through network slimming. In: ICCV

\bibitem[{Nvidia(2018)}]{nvidia}
Nvidia (2018) Nvidia tensor cores.
  \urlprefix\url{https://www.nvidia.com/en-us/data-center/tensorcore/}

\bibitem[{Pham et~al(2018)Pham, Guan, Zoph, Le, and Dean}]{Pham:2018tl}
Pham H, Guan MY, Zoph B, Le QV, Dean J (2018) {Efficient Neural Architecture
  Search via Parameter Sharing}. In: ICML

\bibitem[{Rastegari et~al(2016)Rastegari, Ordonez, Redmon, and
  Farhadi}]{Rastegari:2016tn}
Rastegari M, Ordonez V, Redmon J, Farhadi A (2016) {XNOR-Net - ImageNet
  Classification Using Binary Convolutional Neural Networks.} In: ECCV

\bibitem[{Sandler et~al(2018)Sandler, Howard, Zhu, Zhmoginov, and
  Chen}]{Sandler:2018wy}
Sandler M, Howard A, Zhu M, Zhmoginov A, Chen LC (2018) {MobileNetV2: Inverted
  Residuals and Linear Bottlenecks}. In: CVPR

\bibitem[{Sharma et~al(2018)Sharma, Park, Suda, Lai, Chau, Chandra, and
  Esmaeilzadeh}]{sharma2018bit}
Sharma H, Park J, Suda N, Lai L, Chau B, Chandra V, Esmaeilzadeh H (2018) Bit
  fusion: Bit-level dynamically composable architecture for accelerating deep
  neural network. In: ISCA

\bibitem[{Umuroglu et~al(2018)Umuroglu, Rasnayake, and
  Sjalander}]{umuroglu2018bismo}
Umuroglu Y, Rasnayake L, Sjalander M (2018) Bismo: A scalable bit-serial matrix
  multiplication overlay for reconfigurable computing. In: FPL

\bibitem[{Williams et~al(2009)Williams, Waterman, and
  Patterson}]{williams2009roofline}
Williams S, Waterman A, Patterson D (2009) Roofline: an insightful visual
  performance model for multicore architectures. Communications of the ACM
  52(4):65--76

\bibitem[{Xilinx(2018{\natexlab{a}})}]{vu9p}
Xilinx (2018{\natexlab{a}}) Ultrascale architecture and product data sheet:
  Overview.
  \urlprefix\url{https://www.xilinx.com/support/documentation/data_sheets/ds890-ultrascale-overview.pdf}

\bibitem[{Xilinx(2018{\natexlab{b}})}]{zync7020}
Xilinx (2018{\natexlab{b}}) Zynq-7000 soc data sheet: Overview.
  \urlprefix\url{https://www.xilinx.com/support/documentation/data_sheets/ds190-Zynq-7000-Overview.pdf}

\bibitem[{Yang et~al(2016)Yang, Chen, and Sze}]{yang2016designing}
Yang TJ, Chen YH, Sze V (2016) Designing energy-efficient convolutional neural
  networks using energy-aware pruning. arXiv

\bibitem[{Yang et~al(2018)Yang, Howard, Chen, Zhang, Go, Sandler, Sze, and
  Adam}]{yang2018netadapt}
Yang TJ, Howard A, Chen B, Zhang X, Go A, Sandler M, Sze V, Adam H (2018)
  Netadapt: Platform-aware neural network adaptation for mobile applications.
  In: ECCV

\bibitem[{Zhou et~al(2018)Zhou, Yao, Wang, and Chen}]{zhou2018explicit}
Zhou A, Yao A, Wang K, Chen Y (2018) Explicit loss-error-aware quantization for
  low-bit deep neural networks. In: Proceedings of the IEEE Conference on
  Computer Vision and Pattern Recognition, pp 9426--9435

\bibitem[{Zhou et~al(2016)Zhou, Ni, Zhou, Wen, Wu, and Zou}]{Zhou:2016wh}
Zhou S, Ni Z, Zhou X, Wen H, Wu Y, Zou Y (2016) {DoReFa-Net - Training Low
  Bitwidth Convolutional Neural Networks with Low Bitwidth Gradients.} arXiv

\bibitem[{Zhu et~al(2017)Zhu, Han, Mao, and Dally}]{Zhu:2017wy}
Zhu C, Han S, Mao H, Dally W (2017) {Trained Ternary Quantization}. In: ICLR

\bibitem[{Zoph and Le(2017)}]{Zoph:2017uo}
Zoph B, Le QV (2017) {Neural Architecture Search with Reinforcement Learning}.
  In: ICLR

\end{thebibliography}

\end{document}